\documentclass[10pt,twocolumn,letterpaper]{article}

\usepackage{iccv}
\usepackage{times}
\usepackage{epsfig}
\usepackage{graphicx}
\usepackage{amsmath}
\usepackage{amssymb}
\usepackage{booktabs}
\usepackage{bbding}
\usepackage{subfigure}
\usepackage{diagbox}
\usepackage{setspace}
\usepackage{float}
\usepackage{booktabs}
\usepackage{textcomp}

\usepackage[pagebackref=true,breaklinks=true,letterpaper=true,colorlinks,bookmarks=false]{hyperref}

\iccvfinalcopy 

\def\iccvPaperID{22} 

\ificcvfinal\fi
\begin{document}
\hyphenpenalty=400

\title{S$^3$FD: Single Shot Scale-invariant Face Detector}

\author{Shifeng Zhang\quad Xiangyu Zhu\quad Zhen Lei\thanks{Corresponding author}\quad\  Hailin Shi\quad Xiaobo Wang\quad Stan Z. Li\\
CBSR \& NLPR, Institute of Automation, Chinese Academy of Sciences, Beijing, China\\
University of Chinese Academy of Sciences, Beijing, China\\
{\tt\small \{shifeng.zhang,xiangyu.zhu,zlei,hailin.shi,xiaobo.wang,szli\}@nlpr.ia.ac.cn}
}

\maketitle
\thispagestyle{empty}

\begin{abstract}
This paper presents a real-time face detector, named Single Shot Scale-invariant Face Detector (S$^3$FD), which performs superiorly on various scales of faces with a single deep neural network, especially for small faces. Specifically, we try to solve the common problem that anchor-based detectors deteriorate dramatically as the objects become smaller. We make contributions in the following three aspects: 1) proposing a scale-equitable face detection framework to handle different scales of faces well. We tile anchors on a wide range of layers to ensure that all scales of faces have enough features for detection. Besides, we design anchor scales based on the effective receptive field and a proposed equal proportion interval principle; 2) improving the recall rate of small faces by a scale compensation anchor matching strategy; 3) reducing the false positive rate of small faces via a max-out background label. As a consequence, our method achieves state-of-the-art detection performance on all the common face detection benchmarks, including the AFW, PASCAL face, FDDB and WIDER FACE datasets, and can run at $36$ FPS on a Nvidia Titan X (Pascal) for VGA-resolution images.
\end{abstract}

\setlength{\parskip}{-0.6\baselineskip}
\section{Introduction}
\setlength{\parskip}{-0.3\baselineskip}
Face detection is the key step of many subsequent face-related applications, such as face alignment~\cite{xiong2013supervised,zhu2016face}, face recognition~\cite{parkhi2015deep,schroff2015facenet,zhu2015high}, face verification~\cite{sun2014deep,taigman2014deepface} and face tracking~\cite{kim2008face}, etc. It has been well developed over the past few decades. Following the pioneering work of Viola-Jones face detector~\cite{viola2004robust}, most of early works focus on designing robust features and training effective classifiers. But these approaches depend on non-robust hand-crafted features and optimize each component separately, making the face detection pipeline sub-optimal.
\setlength{\parskip}{-0.0\baselineskip}

\begin{figure}[htbp]
\centering
\subfigure[]{
\label{fig:r-a1} 
\includegraphics[width=4cm]{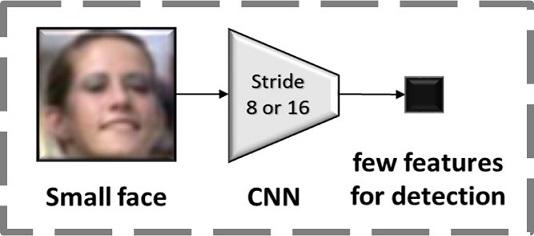}}
\subfigure[]{
\label{fig:r-a2} 
\includegraphics[width=4cm]{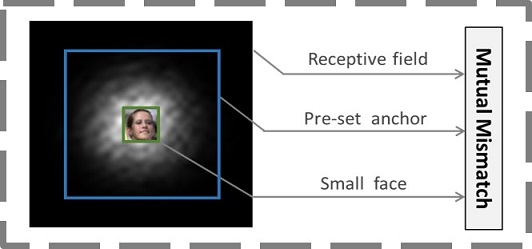}}
\subfigure[]{
\label{fig:r-b} 
\includegraphics[width=4cm]{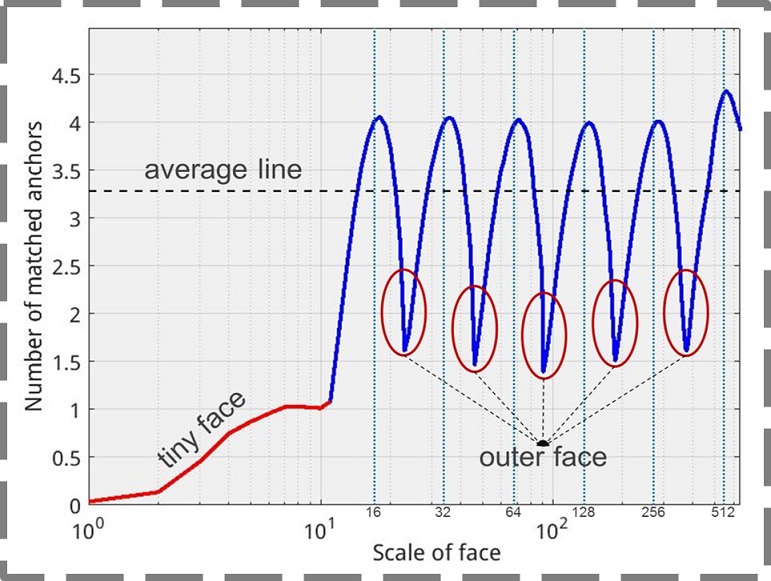}}
\subfigure[]{
\label{fig:r-c} 
\includegraphics[width=4cm]{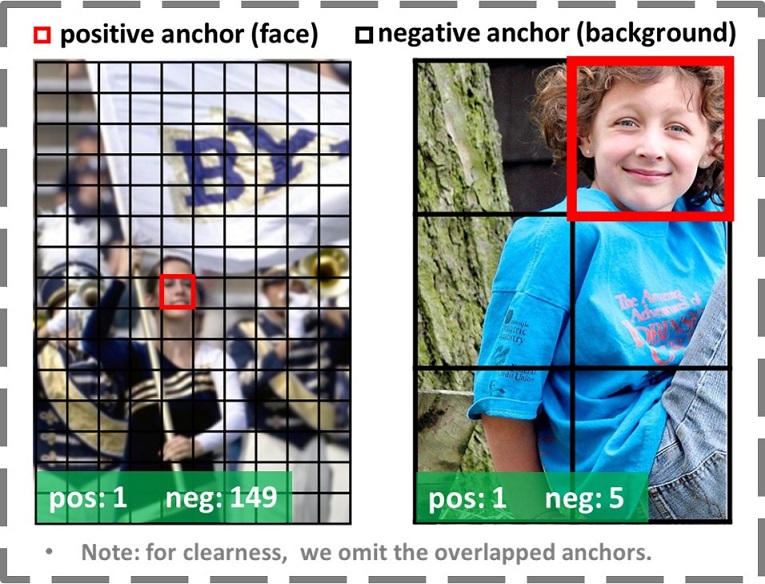}}
\caption{Reasons behind the problem of anchor-based methods. {\bf (a) Few features:} Small faces have few features at detection layer. {\bf (b) Mismatch:} Anchor scale mismatches receptive field and both are too large to fit small face. {\bf (c) Anchor matching strategy:} The figure demonstrates the number of matched anchors at different face scales under current anchor matching method. It reflects that tiny and outer faces match too little anchors. {\bf (d) Background from small anchors:} The two figures have the same resolution. The left one tiles small anchors to detect the small face and the right one tiles big anchors to detect the big face. Small anchors bring about plenty of negative anchors on the background. }
\label{fig:reason} 
\end{figure}

In recent years, convolutional neural network (CNN) has achieved remarkable successes, ranging from image classification~\cite{he2016deep,simonyan2014very,szegedy2016rethinking} to object detection~\cite{girshick2015fast,li2016r,liu2016ssd,redmon2016you,ren2015faster}, which also inspires face detection. On the one hand, many works~\cite{li2015convolutional,ohn2017boost,yang2015convolutional,yang2015facial,zhang2016joint} have applied CNN as the feature extractor in the traditional face detection framewrok. On the other hand, face detection is regarded as a special case of generic object detection and lots of methods~\cite{chen2016supervised,jiang2016face,sun2017face,wan2016bootstrapping,zhu2016cms} have inherited valid techniques from generic object detection method~\cite{ren2015faster}. Following the latter route, we improve the anchor-based generic object detection frameworks and propose a state-of-the-art face detector. 

Anchor-based object detection methods~\cite{liu2016ssd,ren2015faster} detect objects by classifying and regressing a series of pre-set anchors, which are generated by regularly tiling a collection of boxes with different scales and aspect ratios on the image. These anchors are associated with one~\cite{ren2015faster} or several~\cite{liu2016ssd} convolutional layers, whose spatial size and stride size determine the position and interval of the anchors, respectively. The anchor-associated layers are convolved to classify and align the corresponding anchors. Comparing with other methods, anchor-based detection methods are more robust in complicated scenes and their speed is invariant to object numbers. However, as indicated in~\cite{huang2016speed}, {\bf the performance of anchor-based detectors drop dramatically as the objects becoming smaller}. In order to present a scale-invariant anchor-based face detector, we comprehensively analyze the reasons behind the above problem as follows:

{\bf Biased framework.} The anchor-based detection frameworks tend to miss small and medium faces. Firstly, the stride size of the lowest anchor-associated layer is too large (\textit{e.g.}, $8$ pixels in~\cite{liu2016ssd} and $16$ pixels in~\cite{ren2015faster}), therefore small and medium faces have been highly squeezed on these layers and have few features for detection, see Fig.~\ref{fig:r-a1}. Secondly, small face, anchor scale and receptive field are mutual mismatch: anchor scale mismatches receptive field and both are too large to fit small face, see Fig.~\ref{fig:r-a2}. To address the above problems, we propose a scale-equitable face detection framework. We tile anchors on a wide range of layers whose stride size vary from $4$ to $128$ pixels, which guarantees that various scales of faces have enough features for detection. Besides, we design anchors with scales from $16$ to $512$ pixels over different layers according to the effective receptive field~\cite{luo2016understanding} and a new equal-proportion interval principle, which ensures that anchors at different layers match their corresponding effective receptive field and different scales of anchors evenly distribute on the image.

{\bf Anchor matching strategy.} In the anchor-based detection frameworks, anchor scales are discrete (\textit{i.e.}, $16,32,64,128,256,512$ in our method) but face scale is continuous. Consequently, those faces whose scale distribute away from anchor scales can not match enough anchors, such as tiny and outer face in Fig.~\ref{fig:r-b}, leading to their low recall rate. To improve the recall rate of these ignored faces, we propose a scale compensation anchor matching strategy with two stages. The first stage follows current anchor matching method but adjusts a more reasonable threshold. The second stage ensures that every scale of faces match enough anchors through scale compensation.

{\bf Background from small anchors.} To detect small faces well, plenty of small anchors have to be densely tiled on the image. As illustrated in Fig.~\ref{fig:r-c}, these small anchors lead to a sharp increase in the number of negative anchors on the background, bringing about many false positive faces. For example, in our scale-equitable framework, over $75\%$ of negative anchors come from the lowest conv$3\_3$ layer, which is used to detect small faces. In this paper, we propose a max-out background label for the lowest detection layer to reduce the false positive rate of small faces.

For clarity, the main contributions of this paper can be summarized as:
\setlength{\parskip}{-0.8\baselineskip}
\begin{itemize}
\setlength{\itemsep}{0pt}
\setlength{\parsep}{0pt}
\setlength{\parskip}{0pt}
\item Proposing a scale-equitable face detection framework with a wide range of anchor-associated layers and a series of reasonable anchor scales so as to handle different scales of faces well.
\item Presenting a scale compensation anchor matching strategy to improve the recall rate of small faces.
\item Introducing a max-out background label to reduce the high false positive rate of small faces.
\item Achieving state-of-the-art results on AFW, PASCAL face, FDDB and WIDER FACE with real-time speed.
\end{itemize}
\setlength{\parskip}{0\baselineskip}

\section{Related work} \label{2}

Face detection has attracted extensive research attention in past decades. The milestone work of Viola-Jones
~\cite{viola2004robust} uses Haar feature and AdaBoost to train a cascade of face/non-face classifiers that achieves a good accuracy with real-time efficiency. After that, lots of works have focused on improving the performance with more sophisticated hand-crafted features~\cite{liao2016fast,lowe2004distinctive,yang2014aggregate,Zhu2006Fast} and more powerful classifiers~\cite{brubaker2008design,pham2007fast}. Besides the cascade structure,~\cite{mathias2014face,yan2014fastest,zhu2012face} introduce deformable part models (DPM) into face detection tasks and achieve remarkable performance. However, these methods highly depend on the robustness of hand-crafted features and optimize each component separately, making face detection pipeline sub-optimal.

Recent years have witnessed the advance of CNN-based face detectors. CascadeCNN~\cite{li2015convolutional} develops a cascade architecture built on CNNs with powerful discriminative capability and high performance. Qin et al.~\cite{qin2016joint} proposes to jointly train CascadeCNN to realize end-to-end optimization. Faceness~\cite{yang2015facial} trains a series of CNNs for facial attribute recognition to detect partially occluded faces. MTCNN~\cite{zhang2016joint} proposes to jointly solve face detection and alignment using several multi-task CNNs. UnitBox~\cite{yu2016unitbox} introduces a new intersection-over-union loss function.

\begin{figure*}[htbp]
\centering
\includegraphics[width=16cm]{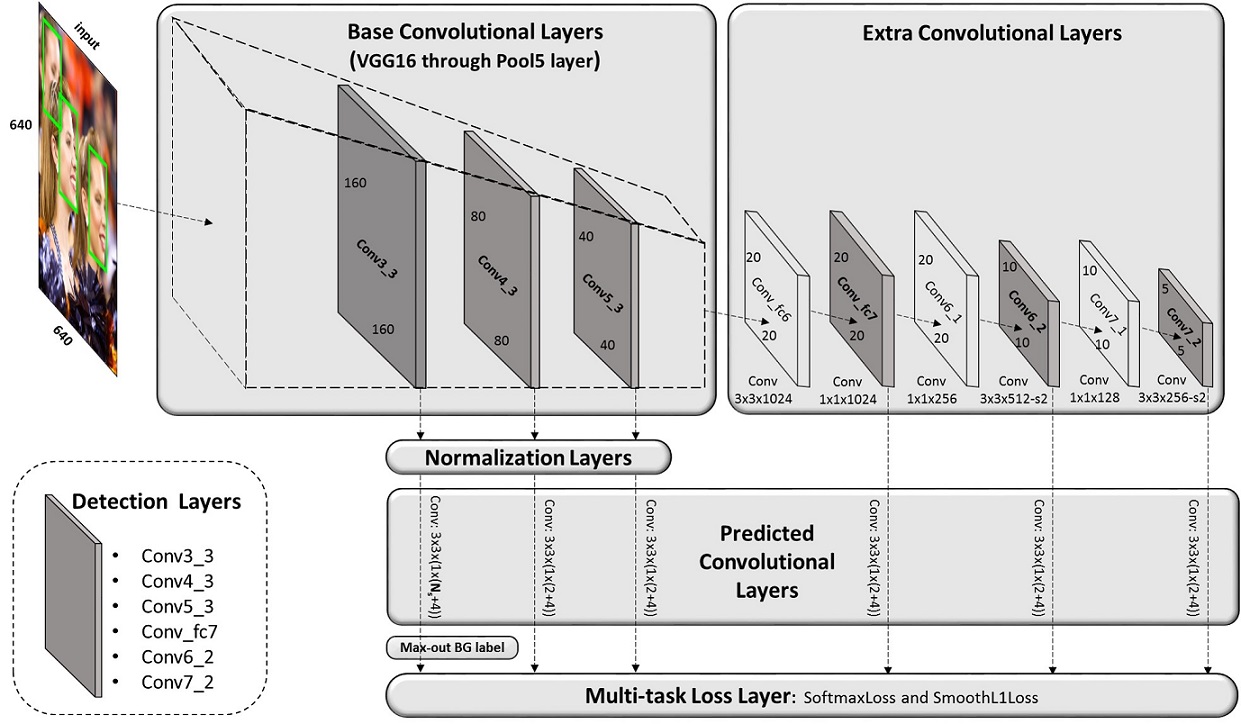}
\caption{Architecture of Single Shot Scale-invariant Face Detector (S$^3$FD). It consists of {\bf Base Convolutional Layers}, {\bf Extra Convolutional Layers}, {\bf Detection Convolutional Layers}, {\bf Normalization Layers}, {\bf Predicted Convolutional Layers} and {\bf Multi-task Loss Layer}.} \label{fig:SFD}
\end{figure*}

Additionally, face detection has inherited some achievements from generic object detection tasks. Jiang et al.~\cite{jiang2016face} applies Faster R-CNN in face detection and achieves promising results. CMS-RCNN~\cite{zhu2016cms} uses Faster R-CNN in face detection with body contextual information. Convnet~\cite{li2016face} integrates CNN with 3D face model in an end-to-end multi-task learning framework. Wan et al.~\cite{wan2016bootstrapping} combines Faster R-CNN with hard negative mining and achieves significant boosts in face detection performance. STN~\cite{chen2016supervised} proposes a new supervised transformer network and a ROI convolution with RPN for face detection. Sun et al.~\cite{sun2017face} presents several effective strategies to improve Faster RCNN for resolving face detection tasks. In this paper, inspired by the RPN in Faster RCNN~\cite{ren2015faster} and the multi-scale mechanism in SSD~\cite{liu2016ssd}, we develop a state-of-the-art face detector with real-time speed.

\section{Single shot scale-invariant face detector} \label{3}

This section introduces our single shot scale-invariant face detector, including the scale-equitable framework (Sec.~\ref{3.1}), the scale compensation anchor matching strategy (Sec.~\ref{3.2}), the max-out background label (Sec.~\ref{3.3}) and the associated training methodology (Sec.~\ref{3.4}).

\subsection{Scale-equitable framework} \label{3.1}

Our scale-equitable framework is based on the anchor-based detection framework, such as RPN~\cite{ren2015faster} and SSD~\cite{liu2016ssd}. Despite its great achievement, the main drawback of the framework is that the performance drops dramatically as the face becomes smaller~\cite{huang2016speed}. To improve the robustness to face scales, we develop a network architecture with a wide range of anchor-associated layers, whose stride size gradually double from 4 to 128 pixels. Hence, our architecture ensures that different scales of faces have adequate features for detection at corresponding anchor-associated layers. After determining the location of anchors, we design the scales of anchors from 16 to 512 pixels based on the effective receptive field and our equal-proportion interval principle. The former guarantees that each scale of anchors matches the corresponding effective receptive field well, and the latter makes different scales of anchors have the same density on the image.

{\bf Constructing architecture.} Our architecture (see Fig.\ref{fig:SFD}) is based on the VGG16~\cite{simonyan2014very} network (truncated before any classification layers) with some auxiliary structures:
\setlength{\parskip}{-0.5\baselineskip}
\begin{itemize}
\setlength{\itemsep}{0pt}
\setlength{\parsep}{0pt}
\setlength{\parskip}{1.5pt}
\item \emph{Base Convolutional Layers}: We keep layers of VGG16 from conv$1\_1$ to pool$5$, and remove all the other layers.
\item \emph{Extra Convolutional Layers}: We convert fc6 and fc7 of VGG16 to convolutional layers by subsampling their parameters~\cite{chen2014semantic}, then add extra convolutional layers behind them. These layers decrease in size progressively and form the multi-scale feature maps. 
\item \emph{Detection Convolutional Layers}: We select {\bf conv$\bf 3\_3$, conv$\bf 4\_3$, conv$\bf 5\_3$, conv$\_$fc7, conv$\bf 6\_2$} and {\bf conv$\bf 7\_2$} as the detection layers, which are associated with different scales of anchor to predict detections.
\item \emph{Normalization Layers}: Comparing to other detection layers, conv$3\_3$, conv$4\_3$ and conv$5\_3$ have different feature scales. Hence we use L2 normalization~\cite{liu2015parsenet} to rescale their norm to $10$, $8$ and $5$ respectively, then learn the scale during the back propagation.
\item \emph{Predicted Convolutional Layers}: Each detection layer is followed by a \emph{p}$\times3\times3\times$\emph{q} convolutional layer, where \emph{p} and \emph{q} are the channel number of input and output, and $3\times3$ is the kernel size. For each anchor, we predict $4$ offsets relative to its coordinates and $N_s$ scores for classification, where $N_s$ = $N_m+1$ ($N_m$ is the max-out background label) for conv$3\_3$ detection layer and $N_s = 2$ for other detection layers. 
\item \emph{Multi-task Loss Layer}: We use softmax loss for classification and smooth L1 loss for regression.
\end{itemize}
\setlength{\parskip}{0\baselineskip}

\begin{table}[htbp]
\centering
\begin{tabular}{cccc}
\toprule[2pt]
{\bf Detection Layer} & {\bf Stride} & {\bf Anchor} & {\bf RF} \\
\midrule[1pt]
conv$3\_3$ & 4 & 16 & 48 \\
conv$4\_3$ & 8 & 32 & 108 \\
conv$5\_3$ & 16 & 64 & 228 \\
conv$\_$fc7 & 32 & 128 & 340 \\
conv$6\_2$ & 64 & 256 & 468 \\
conv$7\_2$ & 128 & 512 & 724 \\
\bottomrule[2pt]
\end{tabular}
\hspace{1in}
\caption{The stride size, anchor scale and receptive field (RF) of the six detection layers. The receptive field here is related to $3\times3$ units on the detection layer, since it is followed by a $3\times3$ predicted convolutional layer to predict detections.}\label{tab:anchor}
\end{table}

\begin{figure}[htbp]
\centering
\subfigure[]{
\label{fig:rf1} 
\includegraphics[width=2.42cm]{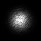}}
\subfigure[]{
\label{fig:rf2} 
\includegraphics[width=2.42cm]{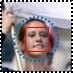}}
\subfigure[]{
\label{fig:anchor} 
\includegraphics[width=3.06cm]{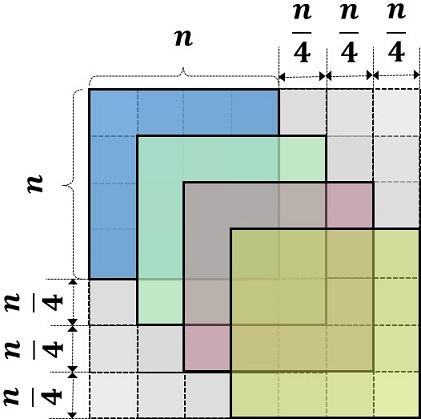}}
\caption{{\bf (a) Effective receptive field:} The whole black box is the theoretical receptive field (TRF) and the white point cloud with Gaussian distribution is the effective receptive field (ERF). ERF only occupies a fraction of TRF. The figure is from~\cite{luo2016understanding}. {\bf (b) A specific example}: In our framework, conv$3\_3$'s TRF is $48\times48$ (the black dotted box) and ERF is the blue dotted circle estimated by (a). Its anchor is $16\times16$ (the red solid line box), which is much smaller than TRF but matches ERF. {\bf (c) Equal-proportion interval principle:} Assuming $n$ is the anchor scale, so $n/4$ is the interval of this scale anchor. $n/4$ also corresponds to the stride size of the layer associated with this anchor. Best viewed in color.}
\label{fig:choice-of-anchor} 
\end{figure}

{\bf Designing scales for anchors.} Each of the six detection layers is associated with a specific scale anchor (\textit{i.e.}, the third column in Tab.~\ref{tab:anchor}) to detect corresponding scale faces. Our anchors are 1:1 aspect ratio (\textit{i.e.}, square anchor), because the bounding box of face is approximately square. As listed in the second and fourth column of Tab.~\ref{tab:anchor}, the stride size and the receptive field of each detection layer are fixed, which are two base points when we design the anchor scales:
\setlength{\parskip}{-0.3\baselineskip}
\begin{itemize}
\setlength{\itemsep}{0pt}
\setlength{\parsep}{0pt}
\setlength{\parskip}{5pt}
\item \emph{Effective receptive field}: As pointed out in~\cite{luo2016understanding}, a unit in the CNN has two types of receptive fields. One is the \emph{theoretical receptive field}, which indicates the input region that can theoretically affect the value of this unit. However, not every pixel in the theoretical receptive field contributes equally to the final output. In general, center pixels have much larger impacts than outer pixels, as shown in Fig.~\ref{fig:rf1}. In other words, only a fraction of the area has effective influence on the output value, which is another type of receptive field, named the \emph{effective receptive field}. According to this theory, the anchor should be significantly smaller than theoretical receptive field in order to match the effective receptive field (see the specific example in Fig.~\ref{fig:rf2}).
\item \emph{Equal-proportion interval principle}: The stride size of a detection layer determines the interval of its anchor on the input image. For example, the stride size of conv$3\_3$ is $4$ pixels and its anchor is $16\times16$, indicating that there is a $16\times16$ anchor for every $4$ pixels on the input image. As shown in the second and third column in Tab.~\ref{tab:anchor}, the scales of our anchors are $4$ times its interval. We call it equal-proportion interval principle (illustrated in Fig.~\ref{fig:anchor}), which guarantees that different scales of anchor have the same density on the image, so that various scales face can approximately match the same number of anchors.
\end{itemize}

Benefits from the scale-equitable framework, our face detector can handle various scales of faces better, especially for small faces.
\setlength{\parskip}{0\baselineskip}

\subsection{Scale compensation anchor matching strategy} \label{3.2}
\begin{figure}[htbp]
\centering
\subfigure[]{
\label{fig:strategy}
\includegraphics[width=4.23cm]{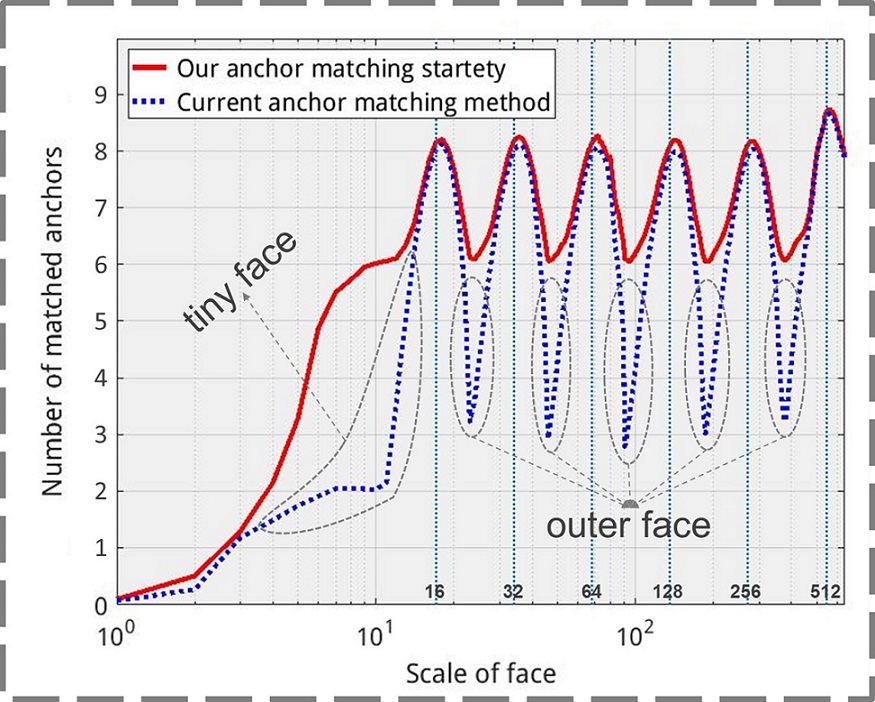}}
\subfigure[]{
\label{fig:max-out}
\includegraphics[width=3.77cm]{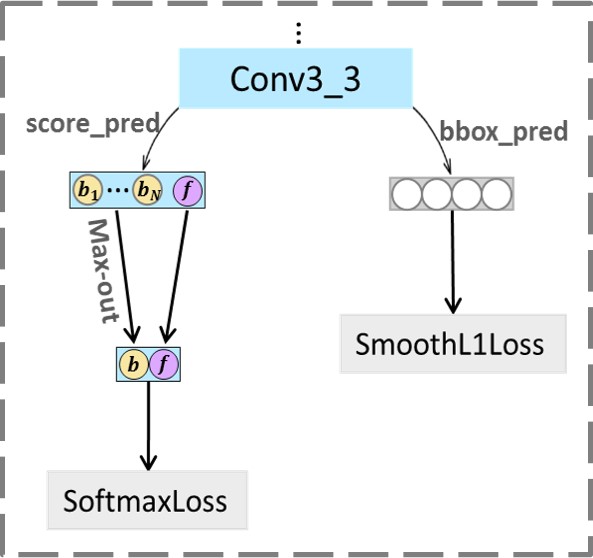}}
\caption{(a) The matched number for different scales of faces are compared between current anchor matching method and our scale compensation anchor matching strategy. (b) The illustration of the max-out background label.}
\end{figure}

During training, we need to determine which anchors correspond to a face bounding box. Current anchor matching method firstly matches each face to the anchors with the best jaccard overlap~\cite{erhan2014scalable} and then matches anchors to any face with jaccard overlap higher than a threshold (usually $0.5$). However, \textbf{anchor scales are discrete while face scales are continuous}, these faces whose scales distribute away from anchor scales can not match enough anchors, leading to their low recall rate. As shown in Fig.~\ref{fig:r-b}, we count the average number of matched anchors for different scales of faces. There are two observations: 1) the average number of matched anchors is about $3$ which is not enough to recall faces with high scores; 2) the number of matched anchors is highly related to the anchor scales. The faces away from anchor scales tend to be ignored, leading to their low recall rate. To solve these problems, we propose a scale compensation anchor matching strategy with two stages:

\setlength{\parskip}{-0.5\baselineskip}
\begin{itemize}
\setlength{\itemsep}{0pt}
\setlength{\parsep}{0pt}
\setlength{\parskip}{1pt}
\item \emph{Stage one}: We follow current anchor matching method but decrease threshold from $0.5$ to $0.35$ in order to increase the average number of matched anchors.
\item \emph{Stage  Two}: After stage one, some faces still do not match enough anchors, such as tiny and outer faces marked with the gray dotted curve in Fig.~\ref{fig:strategy}. We deal with each of these faces as follow: firstly picking out anchors whose jaccard overlap with this face are higher than $0.1$, then sorting them to select top-$N$ as matched anchors of this face. We set $N$ as the average number from stage one.
\end{itemize}

\setlength{\parskip}{-0.5\baselineskip}
As shown in Fig.~\ref{fig:strategy}, our anchor matching strategy greatly increases the matched anchors of tiny and outer faces, which notably improve the recall rate of these faces.
\setlength{\parskip}{0\baselineskip}

\subsection{Max-out background label} \label{3.3}

Anchor-based face detection methods can be regarded as a binary classification problem, which determines if an anchor is face or background. In our method, it is an extremely unbalanced binary classification problem: according to our statistical results, over $99.8\%$ of the pre-set anchors belong to negative anchors (\textit{i.e.}, background) and only a few of anchors are positive anchors (\textit{i.e.}, face). This extreme imbalance is mainly caused by the detection of small faces. Specifically, we have to densely tile plenty of small anchors on the image to detect small faces, which causes a sharp increase in the number of negative anchors. For example, as listed in Tab.~\ref{tab:anchor-e}, a $640\times640$ image has totally $34,125$ anchors, while about $75\%$ of them come from conv$3\_3$ detection layer which is associated with the smallest anchor ($16\times16$). These smallest anchors contribute most to the false positive faces. As a result, improving the detection rate of small faces by tiling small anchors will inevitably lead to the high false positive rate of small faces.
\begin{table}[htbp]
\centering
\begin{tabular}{cccc}
\toprule[2pt]
{\bf Position} & {\bf Scale} & {\bf Number} & {\bf Percentage ($\%$)}\\
\midrule[1pt]
conv3$\_$3 & 16 & 25600 & {\bf 75.02}\\
conv4$\_$3 & 32 & 6400 & 18.76\\
conv5$\_$3 & 64 & 1600 & 4.69\\
conv$\_$fc7 & 128 & 400 & 1.17\\
conv6$\_$2 & 256 & 100 & 0.29\\
conv7$\_$2 & 512 & 25 & 0.07\\
\bottomrule[2pt]
\end{tabular}
\hspace{1in}
\caption{Detailed information about anchors in a 640$\times$640 image.}\label{tab:anchor-e}
\end{table}

To address this issue, we propose to apply a more sophisticated classification strategy on the lowest layer to handle the complicated background from small anchors. We apply the max-out background label for the conv$3\_3$ detection layer. For each of the smallest anchors, we predict $N_m$ scores for background label and then choose the highest as its final score, as illustrated in Fig.~\ref{fig:max-out}. Max-out operation integrates some local optimal solutions into our S$^3$FD model so as to reduce the false positive rate of small faces.

\subsection{Training} \label{3.4}
In this subsection, we introduce the training dataset, data augmentation, loss function, hard negative mining and other implementation details.

{\bf Training dataset and data augmentation.} Our model is trained on $12,880$ images of the WIDER FACE training set with the following data augmentation strategies:
\setlength{\parskip}{-0.5\baselineskip}
\begin{itemize}
\setlength{\itemsep}{0pt}
\setlength{\parsep}{0pt}
\setlength{\parskip}{1pt}
\item Color distort: Applying some photo-metric distortions similar to~\cite{howard2013some}.
\item Random crop: We apply a zoom in operation to generate larger faces since there are too many small faces in the WIDER FACE training set. Specifically, each image is randomly selected from five square patches, which are randomly cropped from the original image: one is the biggest square patch, and the size of the other four square patches range between [$0.3$, $1$] of the short size of the original image. We keep the overlapped part of the face box if its center is in the sampled patch.
\item Horizontal flip: After random cropping, the selected square patch is resized to $640\times640$ and horizontally flipped with probability of $0.5$.
\end{itemize}

\setlength{\parskip}{0\baselineskip}
{\bf Loss function.} We employ the multi-task loss defined in RPN~\cite{DBLP:journals/pami/RenHG017} to jointly optimize model parameters:
\begin{small}
\begin{equation*}\label{1}
\begin{aligned}
\ \ L(\{p_i\},\!\{t_i\})\!&=\!\frac{\lambda}{N\!_{cls}}\!\!\sum_{i}\!L_{cls}(p_i,\!p_i^*)\!+\!\frac{1}{N\!_{reg}}\!\!\sum_{i}\!p_i^*\!L_{reg}(t_i,\!t_i^*),
\end{aligned}
\end{equation*}
\end{small}

\setlength{\parskip}{-0.8\baselineskip}
\noindent where \emph{i} is the index of an anchor and \emph{p$_i$} is the predicted probability that anchor \emph{i} is a face. The ground-truth label \emph{$p_i^*$} is $1$ if the anchor is positive, $0$ otherwise. As defined in~\cite{DBLP:journals/pami/RenHG017}, \emph{t$_i$} is a vector representing the $4$ parameterized coordinates of the predicted bounding box, and \emph{t$_i^*$} is that of the ground-truth box associated with a positive anchor. The classification loss $L_{cls}(p_i,p_i^*)$ is softmax loss over two classes (face vs. background). The regression loss $L_{reg}(t_i,t_i^*)$ is the smooth L1 loss defined in~\cite{girshick2015fast} and \emph{p$_i^*$L$_{reg}$} means the regression loss is activated only for positive anchors and disabled otherwise. The two terms are normalized by \emph{N$_{cls}$} and \emph{N$_{reg}$}, and weighted by a balancing parameter $\lambda$. In our implementation, the \emph{cls} term is normalized by the number of positive and negative anchors, and the \emph{reg} term is normalized by the number of positive anchors. Because of the imbalance between the number of positive and negative anchors, $\lambda$ is used to balance these two loss terms. 
\setlength{\parskip}{0\baselineskip}

{\bf Hard negative mining.} After anchor matching step, most of the anchors are negative, which introduces a significant imbalance between the positive and negative training examples. For faster optimization and stable training, instead of using all or randomly select some negative samples, we sort them by the loss values and pick the top ones so that the ratio between the negatives and positives is at most $3$:$1$. With hard negative mining, we set above background label $N_m=3$, and $\lambda=4$ to balance the loss of classification and regression.

{\bf Other implementation details.} As for the parameter initialization, the base convolutional layers have the same architecture as VGG16 and their parameters are initialized from the pre-trained~\cite{russakovsky2015imagenet} VGG16. The parameters of conv$\_$fc6 and conv$\_$fc7 are initialized by subsampling parameters from fc6 and fc7 of VGG16 and the other additonal layers are randomly initialized with the ``xavier" method~\cite{glorot2010understanding}. We fine-tune the resulting model using SGD with $0.9$ momentum, $0.0005$ weight decay and batch size $32$. The maximum number of iterations is $120k$ and we use $10^{-3}$ learning rate for the first $80k$ iterations, then continue training for $20k$ iterations with $10^{-4}$ and $10^{-5}$. Our method is implemented in Caffe~\cite{jia2014caffe} and the code is available at \url{https://github.com/sfzhang15/SFD}.

\section{Experiments} \label{4}

In this section, we firstly analyze the effectiveness of our scale-equitable framework, scale compensation anchor matching strategy and max-out background label, then evaluate the final model on common face detection benchmarks, finally introduce the inference time.

\subsection{Model analysis} \label{4.1}

We analyze our model on the WIDER FACE validation set by extensive experiments. The WIDER FACE validation set has easy, medium and hard subsets, which roughly correspond to large, medium and small faces, respectively. Hence it is suitable to evaluate our model.

{\bf Baseline.} To evaluate our contributions, we carry out comparative experiments with our baselines. Our S$^3$FD is inspired by RPN~\cite{ren2015faster} and SSD~\cite{liu2016ssd}, so we directly use them to train two face detectors as the baselines, marked as RPN-face and SSD-face, respectively. Different from~\cite{ren2015faster}, the RPN-face tiles six scales of the square anchor ($16$, $32$, $64$, $128$, $256$, $512$) on the conv$5\_3$ layer of VGG16 to make the comparison more substantial. The SSD-face inherits the architecture and anchor-setting of SSD. The remainder is set as the same with our S$^3$FD.

{\bf Ablative Setting.} To understand S$^3$FD better, we conduct ablation experiments to examine how each proposed component affects the final performance. We evaluate the performance of our method under three different settings: (i) \emph{S$^3$FD(F)}: it only uses the scale-equitable framework (\textit{i.e.}, constructed architecture and designed anchors) and ablates another two components; (ii) \emph{S$^3$FD(F+S)}: it applies the scale-equitable framework and the scale compensation anchor matching strategy; (iii) \emph{S$^3$FD(F+S+M)}: it is our complete model, consisting of the scale-equitable framework, the scale compensation anchor matching strategy and the max-out background label. 

\begin{table}[!htbp]
\centering
\begin{tabular}{|l|ccc|}
\hline
\diagbox[height=35pt,width=35mm]{{\bf Methods}}{{\bf mAP(\%)}}{{\bf Subsets}}& Easy & Medium & Hard \\
\hline
RPN-face &91.0&88.2&73.7\\
SSD-face &92.1&89.5&71.6\\
S$^3$FD(F) &92.6&91.6&82.3\\
S$^3$FD(F+S) &93.5&92.0&84.5 \\
S$^3$FD(F+S+M) &{\bf 93.7}&{\bf 92.4}&{\bf 85.2}\\
\hline
\end{tabular}
\hspace{1in}
\caption{The comparative and ablative results of our model on WIDER FACE validation subset. The precision-recall curves of these methods are in the supplementary materials.}
\label{tab:ablation}
\end{table}
From the results listed in Tab.~\ref{tab:ablation}, some promising conclusions can be summed up as follows:

{\bf Scale-equitable framework is crucial.} Comparing with S$^3$FD(F), the only difference with RPN-face and SSD-face are their framework. RPN-face has the same choice of anchors as ours but only tiles on the last convolutional layer of VGG16. Not only its stride size ($16$ pixels) is too large for small faces, but also different scales of anchors have the same receptive field. SSD-face tiles anchors over several convolutional layers, while its smallest stride size ($8$ pixels) and smallest anchors are still slightly large for small faces. Besides, its anchors do not match the effective receptive field. The result of S$^3$FD(F) in Tab.~\ref{tab:ablation} shows that our framework greatly outperforms SSD-face and RPN-face, especially on the hard subsets (rising by $8.6\%$), which mainly consists of small faces. Comparing the results between different subsets, our S$^3$FD(F) handles various scales of faces well, and deteriorates slightly as the faces become smaller, which demonstrates the robustness to face scales.

{\bf Scale compensation anchor matching strategy is better.} 
The comparison between the third and fourth rows in Tab.~\ref{tab:ablation} indicates that our scale compensation anchor matching strategy effectively improves the performance, especially for small faces. The mAP is increased by $0.9\%$, $0.4\%$, $2.2\%$ on easy, medium and hard subset, respectively. The increases mainly come from the higher recall rate of various scales of faces, especially for those faces that are ignored by the current anchor matching method.

{\bf Max-out background label is promising.} The last contribution of S$^3$FD is the max-out background label. It deals with the massive small negative anchors (\textit{i.e.}, background) from the conv$3\_3$ detection layer which is designed to detect small faces. As reported in Tab.~\ref{tab:ablation}, the improvements on easy, medium and hard subsets are $0.2\%$, $0.4\%$, $0.7\%$, respectively. It demonstrates that the effectiveness of the max-out background label is positively related to the difficulty of the input image. Since the harder images will generate the more difficult small backgrounds.

\subsection{Evaluation on benchmark}
We evaluate our S$^3$FD method on all the common face detection benchmarks, including Annotated Faces in the Wild (AFW)\cite{zhu2012face}, PASCAL Face\cite{yan2014face}, Face Detection Data Set and Benchmark (FDDB)\cite{jain2010fddb} and WIDER FACE~\cite{yang2016wider}. Due to the limited space, some qualitative results on these dataset will be shown in the supplementary materials.

\begin{figure}[htbp]
\centering
\includegraphics[width=0.4\textwidth]{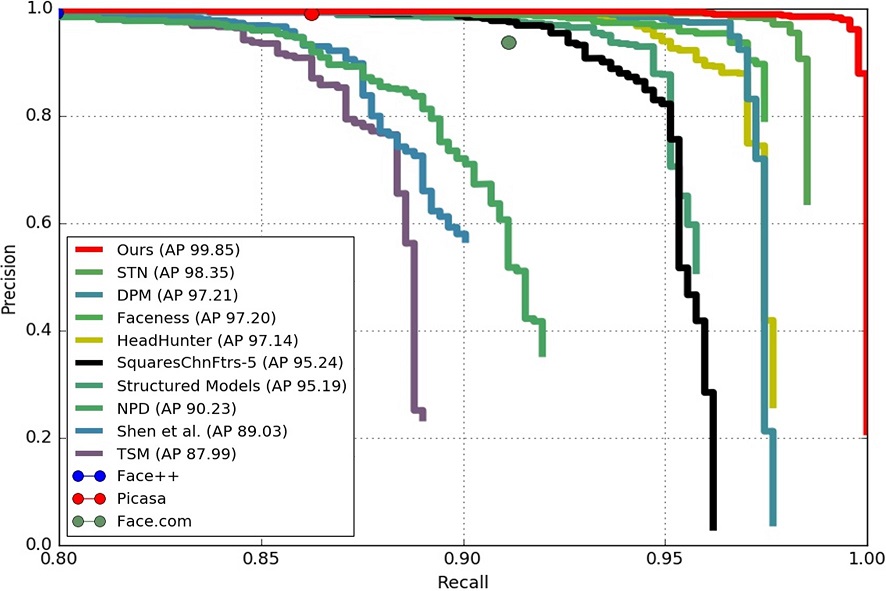}
\caption{Precision-recall curves on AFW dataset.}
\label{fig:AFW}
\end{figure}

{\bf AFW dataset.} It contains $205$ images with $473$ labeled faces. We evaluate our model against the well-known works~\cite{chen2016supervised,liao2016fast,mathias2014face,shen2013detecting,yan2014face,yang2015facial,zhu2012face} and commercial face detectors (\textit{e.g.}, Face.com, Face++ and Picasa). As illustrated in Fig.\ref{fig:AFW}, our S$^3$FD outperforms all others by a large margin. 
\vspace{-0.17cm}

\begin{figure}[htbp]
\centering
\includegraphics[width=0.4\textwidth]{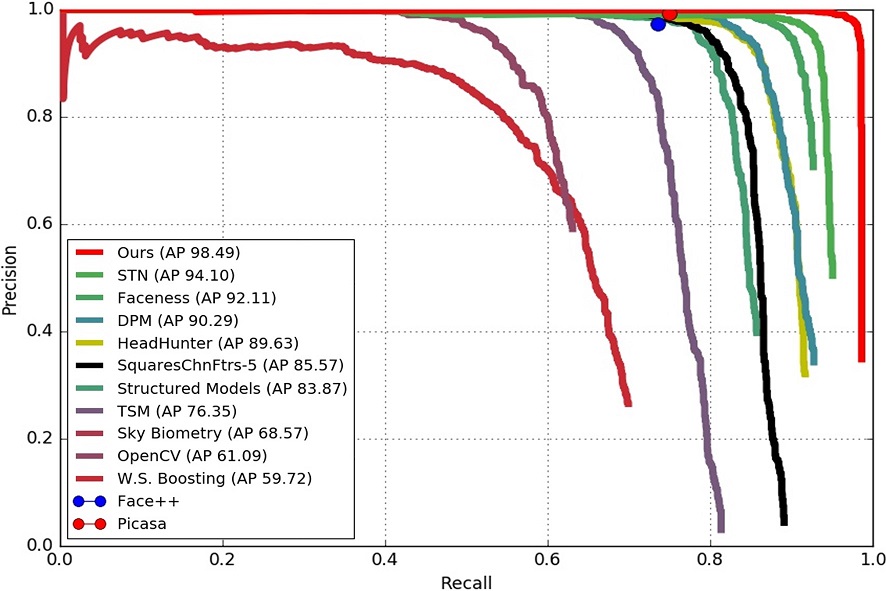}
\caption{Precision-recall curves on PASCAL face dataset.}
\label{fig:PASCAL}
\end{figure}

{\bf PASCAL face dataset.} It has $1,335$ labeled faces in $851$ images with large face appearance and pose variations. It is collected from PASCAL person layout test subset. Fig.\ref{fig:PASCAL} shows the precision-recall curves on this dataset, our method significantly outperforms all other methods~\cite{chen2016supervised,kalal2008weighted,mathias2014face,yan2014face,yang2015facial,zhu2012face} and commercial face detectors (\textit{e.g.}, SkyBiometry, Face++ and Picasa).
\vspace{-0.17cm}

\begin{figure}[htbp]
\centering
\subfigure[Discontinuous ROC curves]{
\label{fig:FDDBd}
\includegraphics[width=0.425\textwidth]{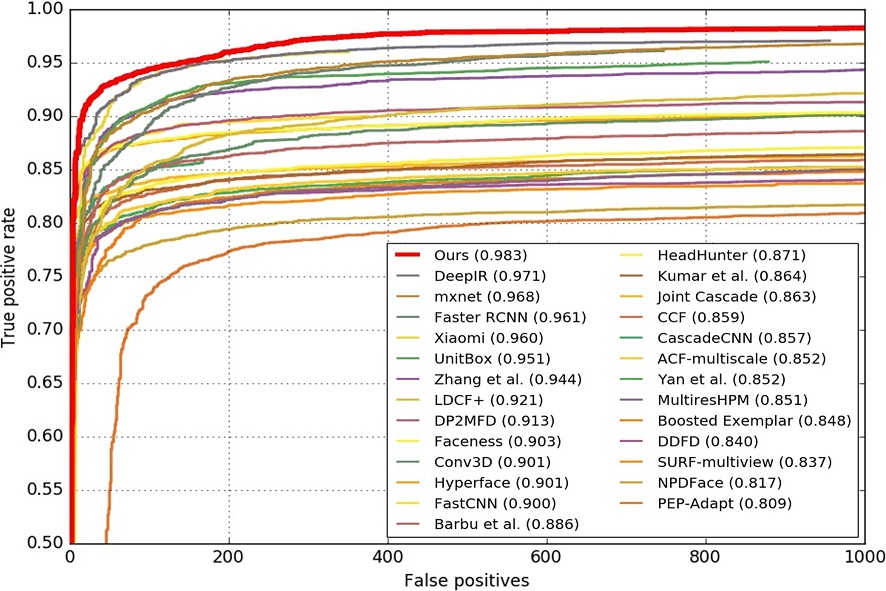}}
\subfigure[Continuous ROC curves]{
\label{fig:FDDBc} 
\includegraphics[width=0.425\textwidth]{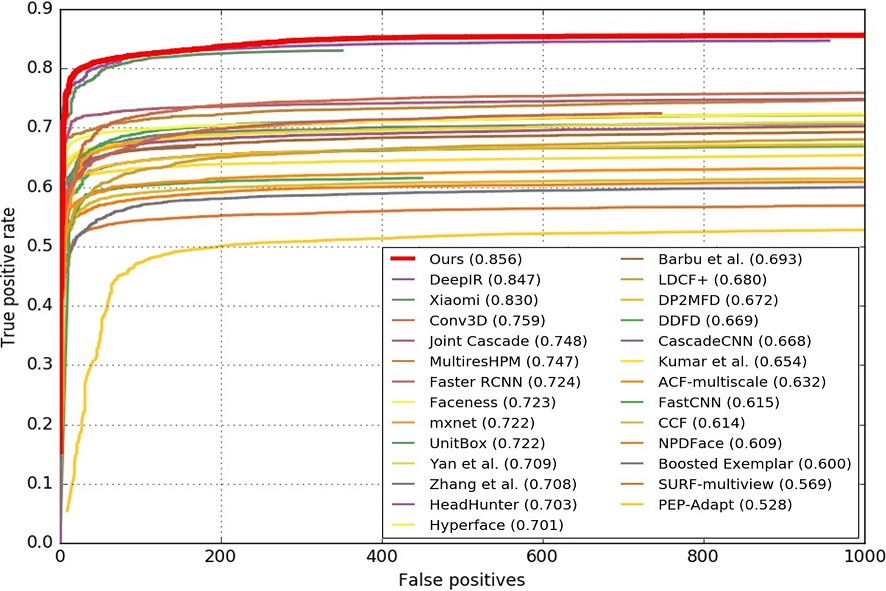}}
\caption{Evaluation on the FDDB dataset.}
\label{fig:Evaluation}
\end{figure}

{\bf FDDB dataset.} It contains $5,171$ faces in $2,845$ images. There are two problems for evaluation: 1) FDDB adopts the bounding ellipse while our S$^3$FD outputs rectangle bounding box. This inconsistency has a great impact on the continuous score, so we train an elliptical regressor to transform our predicted bounding boxes to bounding ellipses. 2) FDDB has lots of unlabelled faces, which results in many false positive faces with high scores. Hence, we manually review the results and add 238 unlabelled faces (annotations will be released later and some examples are shown in the supplementary materials). Finally, we evaluate our face detector on FDDB against the state-of-the-art methods~\cite{barbu2014face,farfade2015multi,ghiasi2015occlusion,jiang2016face,kumar2015visual,li2013probabilistic,li2014efficient,li2013learning,li2016face,liao2016fast,ohn2017boost,ranjan2015deep,ranjan2016hyperface,sun2017face,triantafyllidou2016fast,wan2016bootstrapping,yang2015facial,yu2016unitbox,zhang2016joint}. The results are shown in Fig.~\ref{fig:FDDBd} and Fig.\ref{fig:FDDBc}. Our S$^3$FD achieves the state-of-the-art performance and outperforms all others by a large margin on discontinuous and continuous ROC curves. These results indicate that our S$^3$FD can robustly detect unconstrained faces.
\vspace{-0.17cm}

\makeatletter
\renewcommand\@makefnmark{\hbox{\@textsuperscript{\normalfont\color{red}\@thefnmark}}}
\renewcommand\@makefntext[1]{%
  \parindent 1em\noindent
            \hb@xt@1.8em{%
                \hss\@textsuperscript{\normalfont\@thefnmark}}#1}
\makeatother

\begin{figure*}[htbp]
\centering
\subfigure[Val: Easy]{
\label{fig:ve} 
\includegraphics[width=0.325\textwidth]{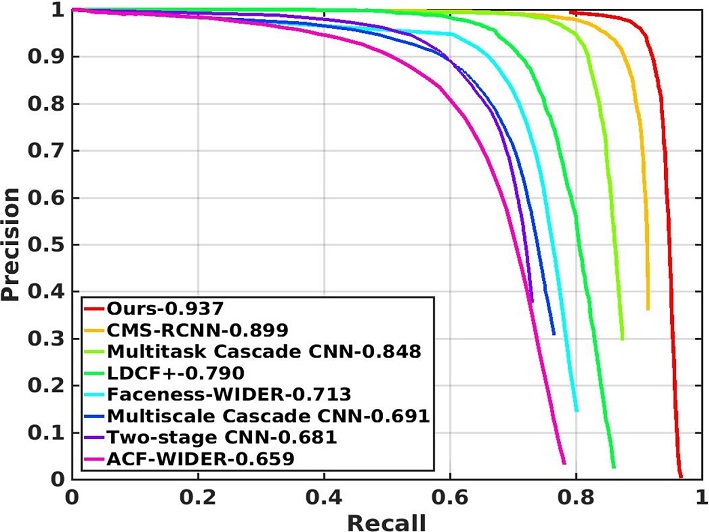}}
\subfigure[Val: Medium]{
\label{fig:vm} 
\includegraphics[width=0.325\textwidth]{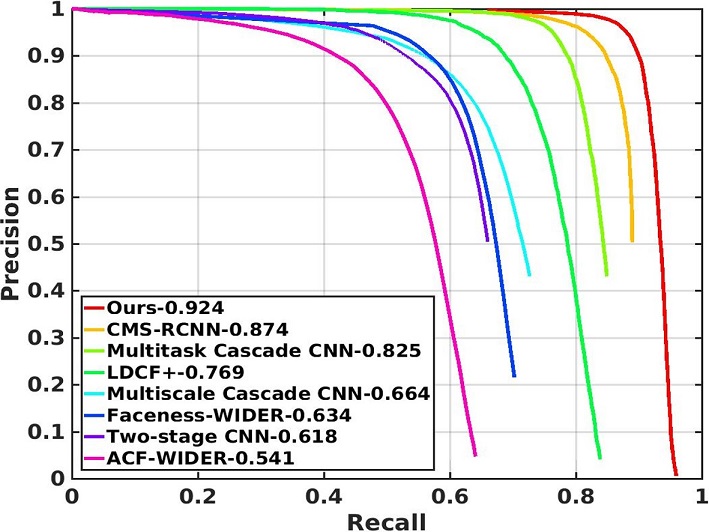}}
\subfigure[Val: Hard]{
\label{fig:vh} 
\includegraphics[width=0.325\textwidth]{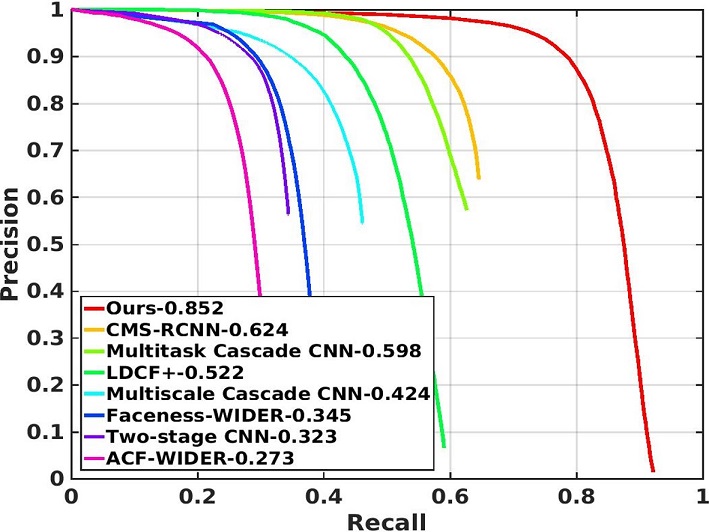}}
\subfigure[Test: Easy]{
\label{fig:te} 
\includegraphics[width=0.325\textwidth]{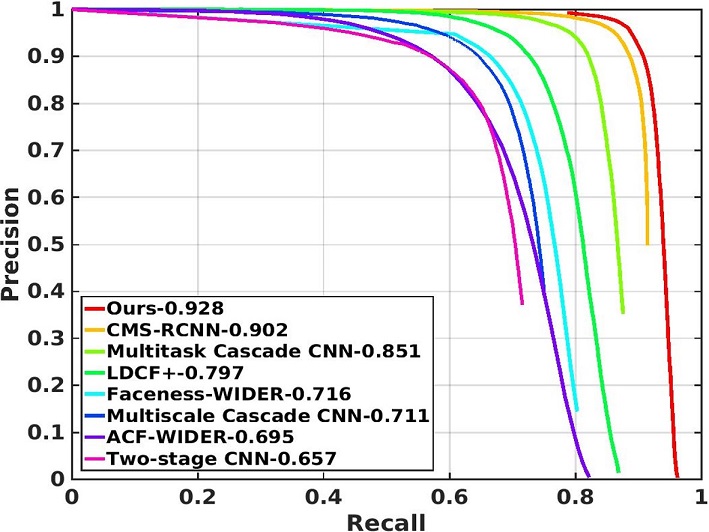}}
\subfigure[Test: Medium]{
\label{fig:tm} 
\includegraphics[width=0.325\textwidth]{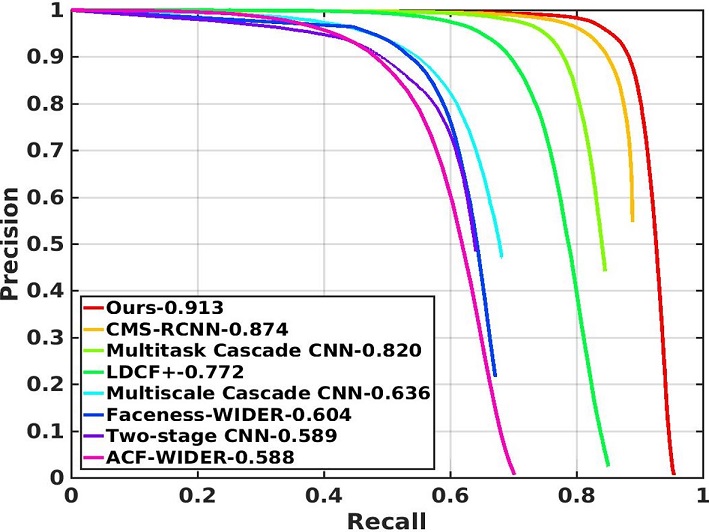}}
\subfigure[Test: Hard]{
\label{fig:th} 
\includegraphics[width=0.325\textwidth]{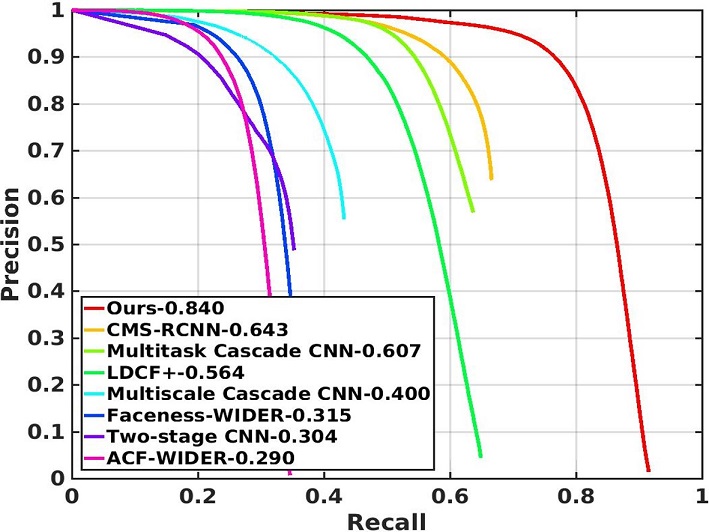}}
\caption{Precision-recall curves on WIDER FACE validation and test sets.\protect\footnotemark[1]}
\label{fig:WIDER_FACE} 
\end{figure*}

{\bf WIDER FACE dataset.} It has $32,203$ images and labels $393,703$ faces with a high degree of variability in scale, pose and occlusion. The database is split into training ($40\%$), validation ($10\%$) and testing ($50\%$) set. Besides, the images are divided into three levels (Easy, Medium and Hard subset) according to the difficulties of the detection. The images and annotations of training and validation set are available online, while the annotations of testing set are not released and the results are sent to the database server for receiving the precision-recall curves. Our S$^3$FD is trained only on the training set and tested on both validation and testing set against recent face detection methods~\cite{ohn2017boost,yang2014aggregate,yang2015facial,yang2016wider,zhang2016joint,zhu2016cms}. The precision-recall curves and mAP values are shown in Fig.~\ref{fig:WIDER_FACE}. Our model outperforms others by a large margin across the three subsets, especially on the hard subset which mainly contains small faces. It achieves the best average precision in all level faces, i.e. $0.937$ (Easy), $0.924$ (Medium) and $0.852$ (Hard) for validation set, and $0.928$ (Easy), $0.913$ (Medium) and $0.840$ (Hard) for testing set.\footnote{Our latest results on WIDER FACE are shown in Fig.~\ref{fig:sWIDER_FACE}} These results not only demonstrate the effectiveness of the proposed method but also strongly show the superiority of the proposed model in detecting small and hard faces.

\subsection{Inference time}
During inference, our method outputs a large number of boxes (\textit{e.g.}, $25,600$ boxes for a VGA-resolution image). To speed up the inference time, we first filter out most boxes by a confidence threshold of $0.05$ and keep the top $400$ boxes before applying NMS, then we perform NMS with jaccard overlap of $0.3$ and keep the top $200$ boxes. We measure the speed using Titan X (Pascal) and cuDNN v5.1 with Intel Xeon E5-2683v3@2.00GHz. For the VGA-resolution image with batch size $1$ using a single GPU, our face detector can run at $36$ FPS and achieve the real-time speed. Besides, about 80$\%$ of the forward time is spent on the VGG16 network, hence using a faster base network could further improve the speed.

\section{Conclusion} \label{5}
This paper introduces a novel face detector by solving the common problem of anchor-based detection methods whose performance decrease sharply as the objects becoming smaller. We analyze the reasons behind this problem, and propose a scale-equitable framework with a wide range of anchor-associated layers and a series of reasonable anchor scales in order to well handle different scales of faces. Besides, we propose the scale compensation anchor matching strategy to improve the recall rate of small faces, and the max-out background label to reduce the false positive rate of small faces. The experiments demonstrate that our three contributions lead S$^3$FD to the state-of-the-art performance on all the common face detection benchmarks, especially for small faces. In our future work, we intend to further improve the classification strategy of background patches. We believe that explicitly dividing the background class into some sub-categories is worthy of further study.

\section*{Acknowledgments} \label{6}
This work was supported by the National Key Research and Development Plan (Grant No.2016YFC0801002), the Chinese National Natural Science Foundation Projects $\#61473291$, $\#61502491$, $\#61572501$, $\#61572536$, $\#61672521$ and AuthenMetric R\&D Funds.

\begin{spacing}{1.0}
{\bibliographystyle{ieee}
\bibliography{egbib}
}
\end{spacing}

\onecolumn
\begin{center}

\ \ 
\vspace{0.3cm}

\title{\Large \textbf{S$^3$FD: Single Shot Scale-invariant Face Detector}}

\vspace{0.25cm}

\title{\Large \textbf{-Supplementary Material-}}

\vspace{0.5cm}

\author{\large Shifeng Zhang\quad Xiangyu Zhu\quad Zhen Lei\quad\  Hailin Shi\quad Xiaobo Wang\quad Stan Z. Li\\
CBSR \& NLPR, Institute of Automation, Chinese Academy of Sciences, Beijing, China\\
University of Chinese Academy of Sciences, Beijing, China\\
{\tt\small \{shifeng.zhang,xiangyu.zhu,zlei,hailin.shi,xiaobo.wang,szli\}@nlpr.ia.ac.cn}
}
\end{center}

\setcounter{section}{0}
\renewcommand\thesection{\Alph{section}} 
\section{Precision-recall curves}
\begin{figure}[H]
\centering
\subfigure[Easy]{
\label{fig:sve} 
\includegraphics[width=0.325\textwidth]{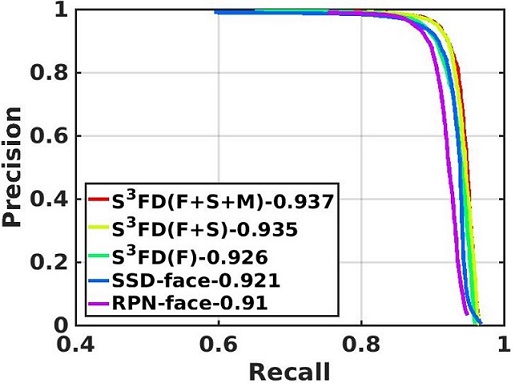}}
\subfigure[Medium]{
\label{fig:svm} 
\includegraphics[width=0.325\textwidth]{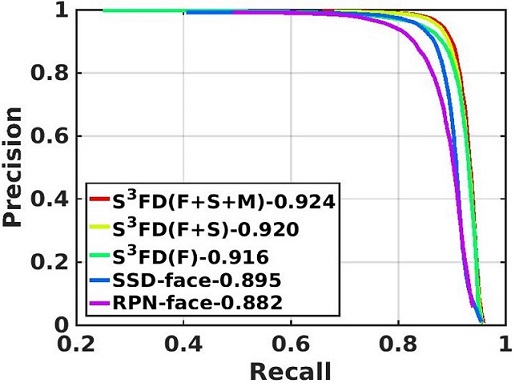}}
\subfigure[Hard]{
\label{fig:svh} 
\includegraphics[width=0.325\textwidth]{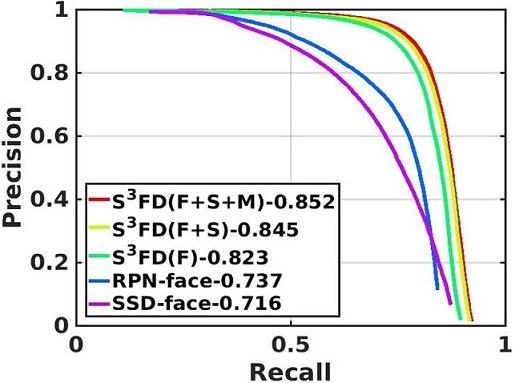}}
\caption{Precision-recall curves on WIDER FACE validation set.}\label{fig:spr}
\end{figure}

In our submitted paper, Tab.~\ref{tab:ablation} in subsection~\ref{4.1} only provides the mAP of RPN-face, SSD-face, S$^3$FD(F), S$^3$FD(F+S) and S$^3$FD(F+S+M). Their precision-recall curves on the WIDER FACE validation set are shown in Fig.~\ref{fig:spr} for details.

\section{Qualitative results}
\begin{figure}[H]
\centering
\includegraphics[width=1.0\textwidth]{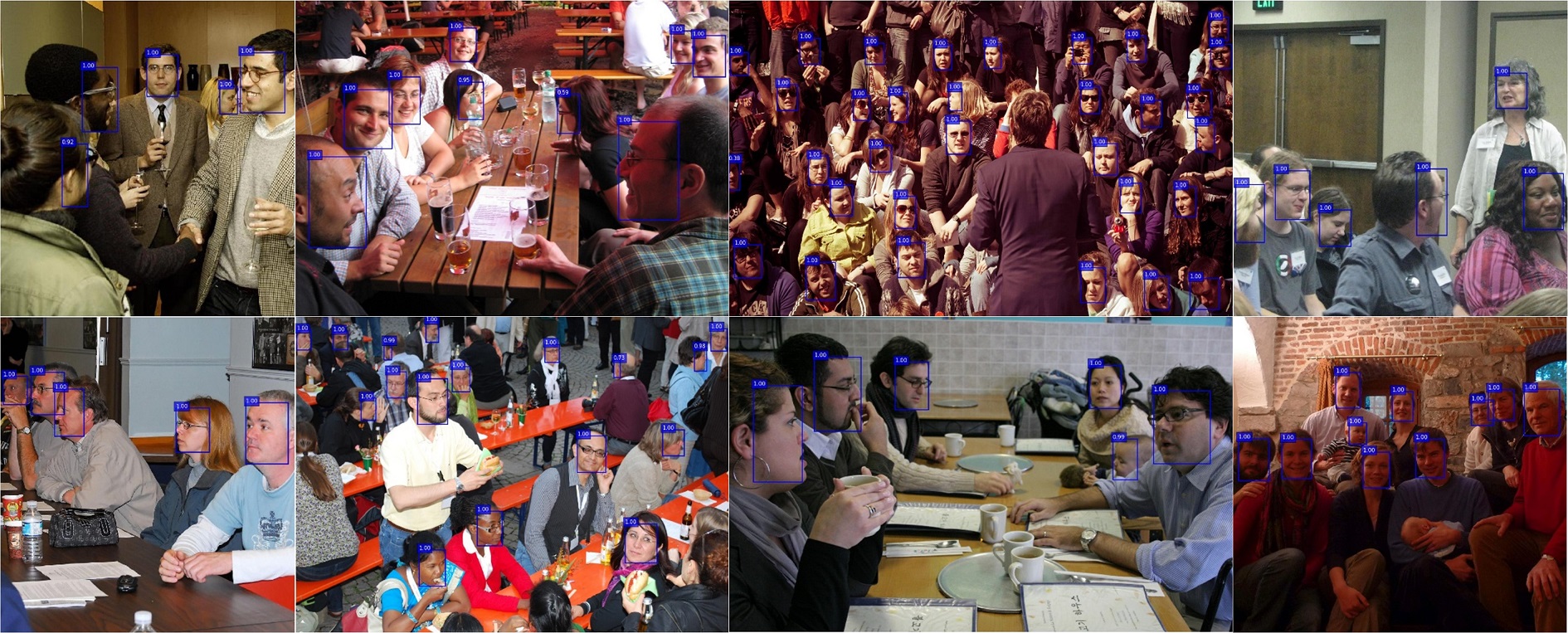}
\caption{Qualitative results on AFW. The faces in these results have a high degree of variability in scale, pose and occlusion. Our S$^3$FD is able to detect these faces with a high confidence, especially for small faces. Please zoom in to see some small detections.}\label{fig:safw}
\end{figure}

\begin{figure}[H]
\centering
\includegraphics[width=1.0\textwidth]{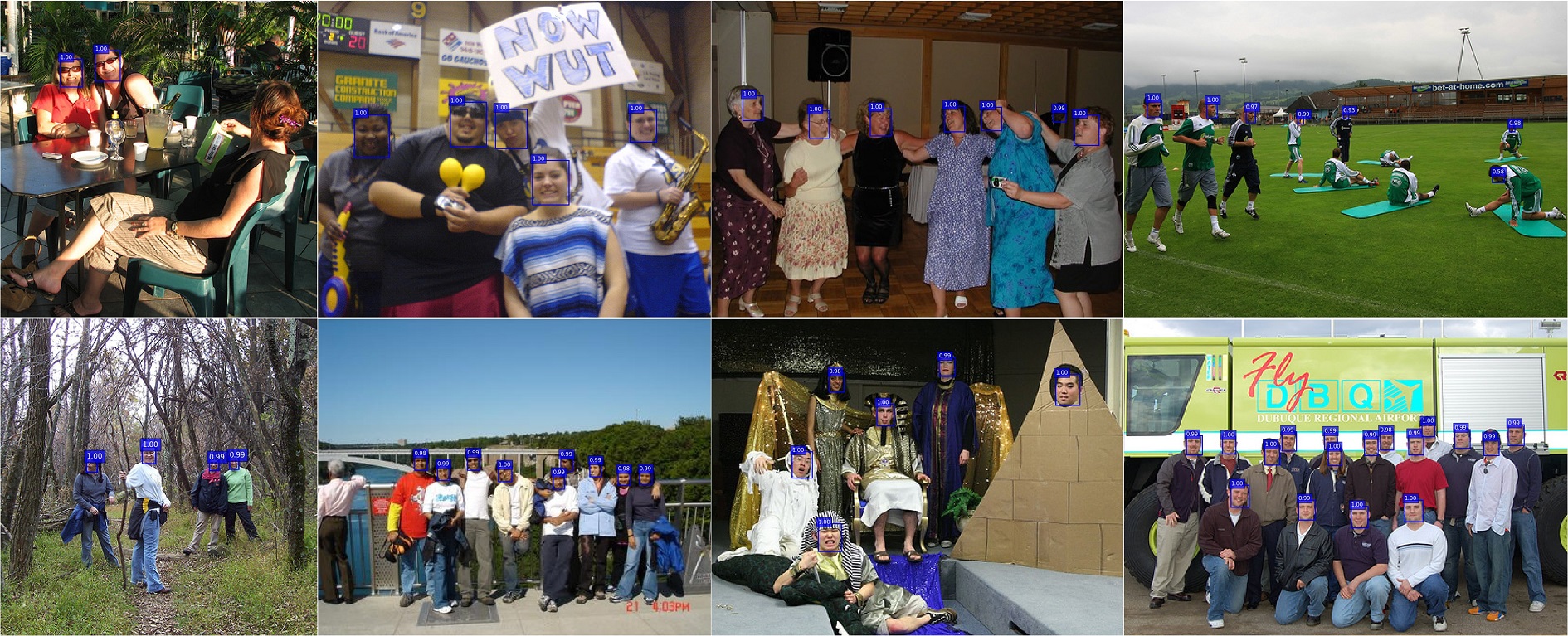}
\caption{Qualitative results on PASCAL face. Most faces in these results are small faces, because the image in PASCAL face has a low resolution. Our S$^3$FD is able to handle small faces well. Please zoom in to see some small detections.}\label{fig:spascal}
\end{figure}

\begin{figure}[H]
\centering
\includegraphics[width=1.0\textwidth]{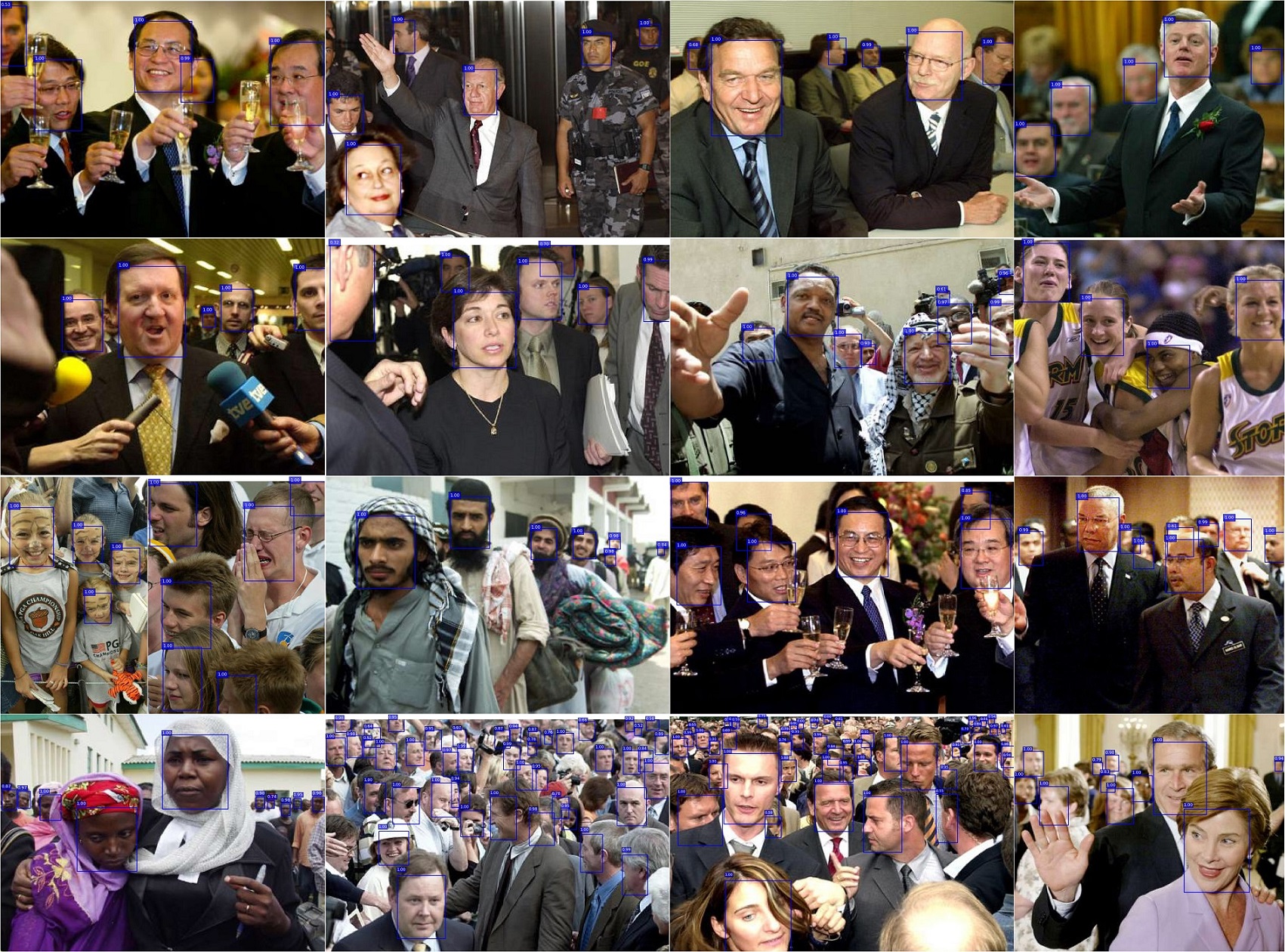}
\caption{Qualitative results on FDDB. These results indicate that our S$^3$FD is robust to large appearance, heavy occlusion, scale variance and heavy blur. Please zoom in to see some small detections.}\label{fig:sfddb}
\end{figure}

\begin{figure}[H]
\centering
\subfigure[Scale attribute. Our S$^3$FD is able to detect faces at a continuous range of scales.]{
\includegraphics[width=1.0\textwidth]{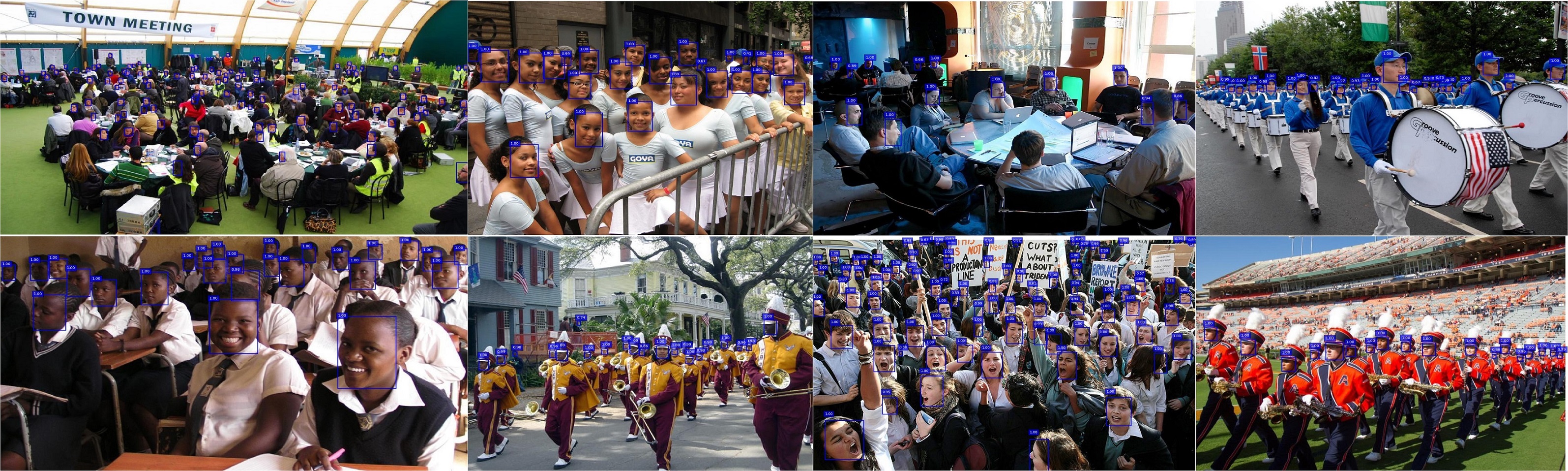}}
\subfigure[Our S$^3$FD is robust to pose, occlusion, expression, makeup, illumination and blur.]{
\includegraphics[width=1.0\textwidth]{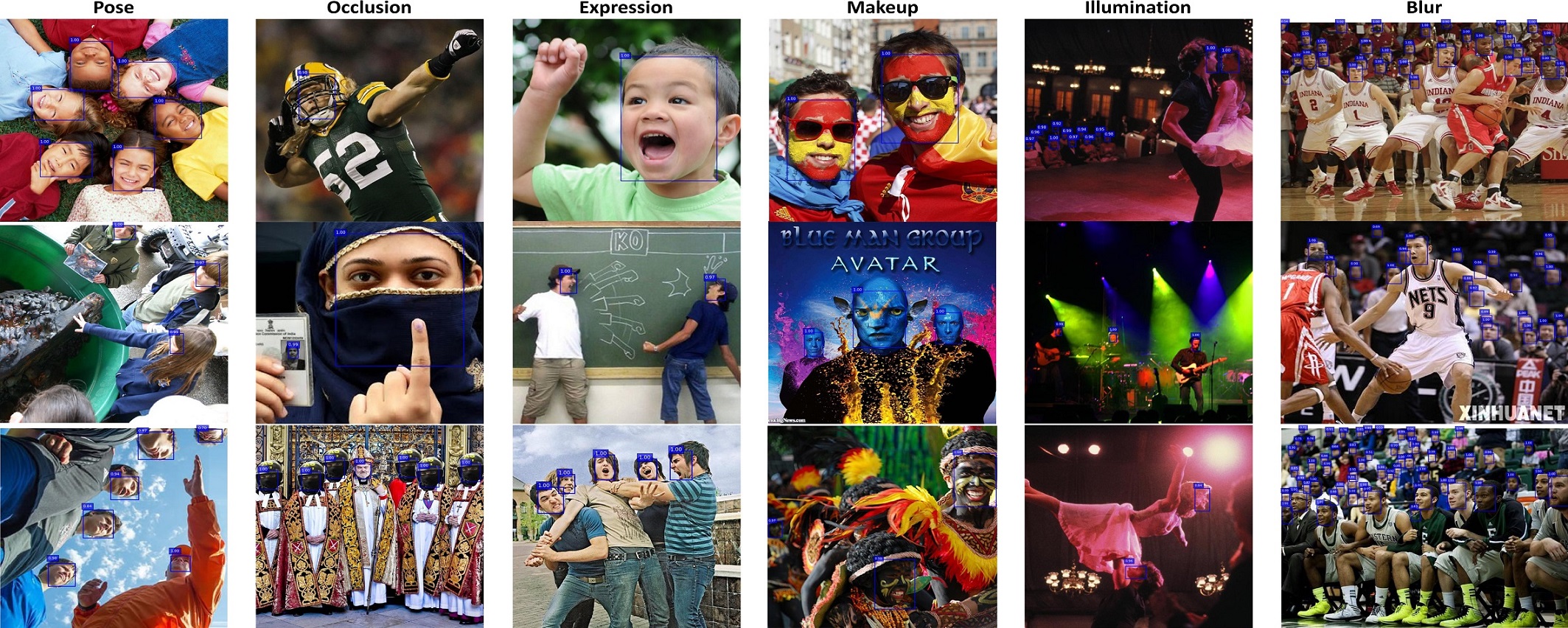}}
\caption{Qualitative results on WIDER FACE. We visualize some examples for each attribute. Please zoom in to see small detections.}\label{fig:swf}
\end{figure}

\begin{figure}[H]
\centering
\includegraphics[width=1.0\textwidth]{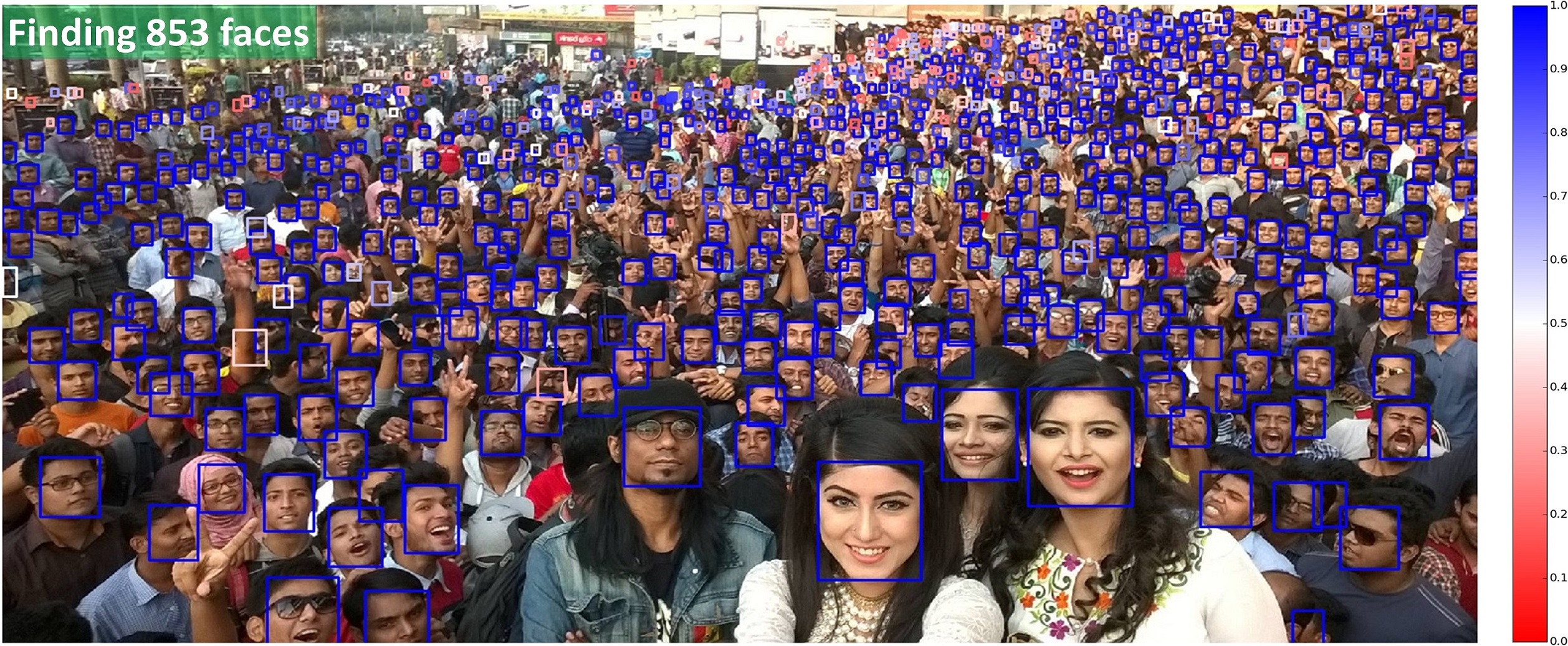}
\caption{Another qualitative result. Our S$^3$FD can find 853 faces out of the reportedly 1000 present in the above image. Detector confidence is given by the colorbar on the right. Please zoom in to see some small detections.}\label{fig:slumia}
\end{figure}

In this section, we demonstrate some qualitative results on common face detection benchmarks, including AFW (Fig.~\ref{fig:safw}), PASCAL face (Fig.~\ref{fig:spascal}), FDDB (Fig.~\ref{fig:sfddb}) and WIDER FACE (Fig.~\ref{fig:swf}). Besides, another impressive result is shown in Fig.~\ref{fig:slumia}.

\section{Examples of manually labelled faces on FDDB}
\begin{figure}[H]
\centering
\subfigure[Profile faces]{
\includegraphics[width=1.0\textwidth]{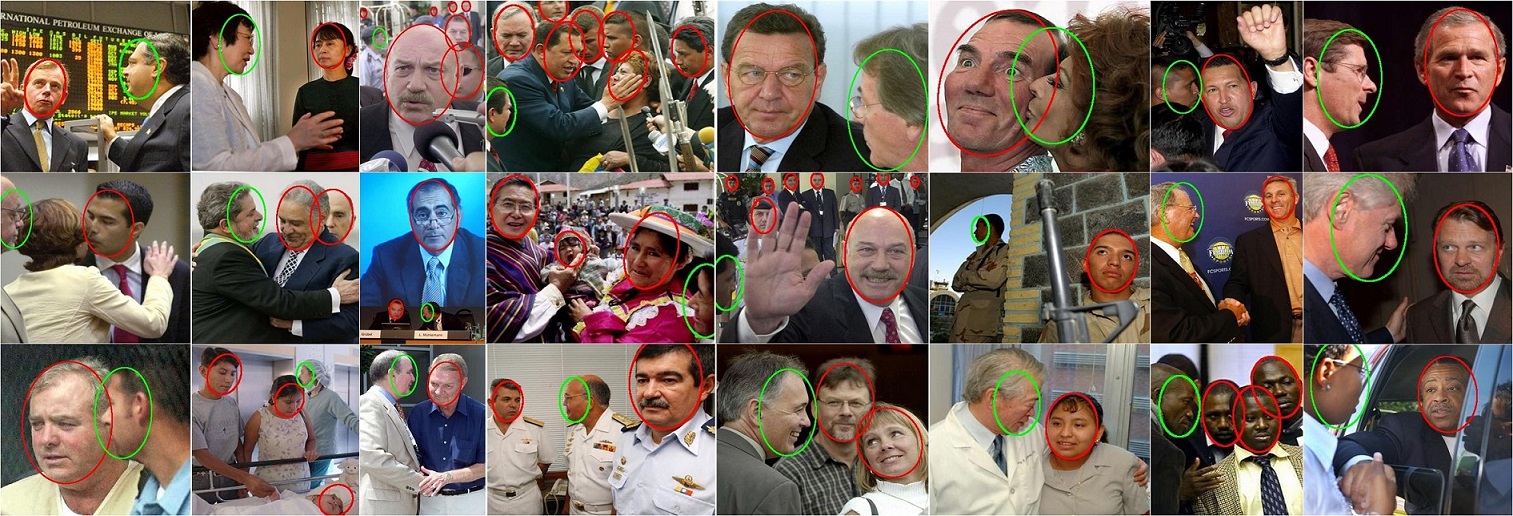}}
\subfigure[Occluded faces]{
\includegraphics[width=1.0\textwidth]{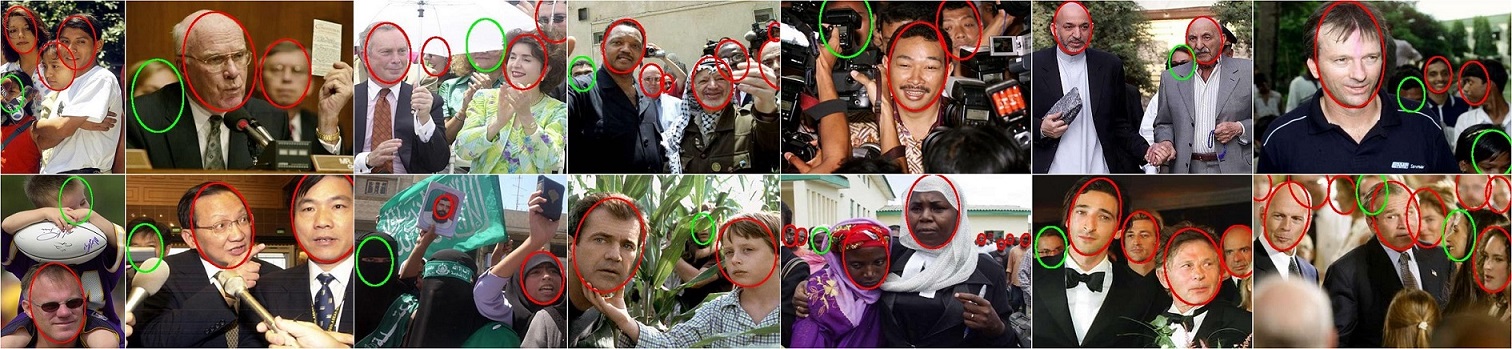}}
\subfigure[Blur faces]{
\includegraphics[width=1.0\textwidth]{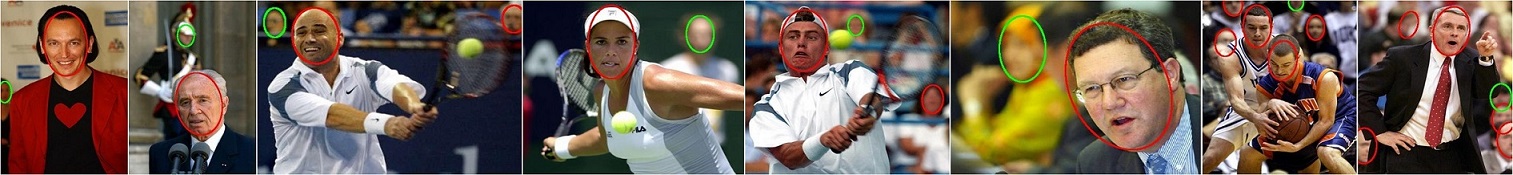}}
\subfigure[Statue faces]{
\includegraphics[width=1.0\textwidth]{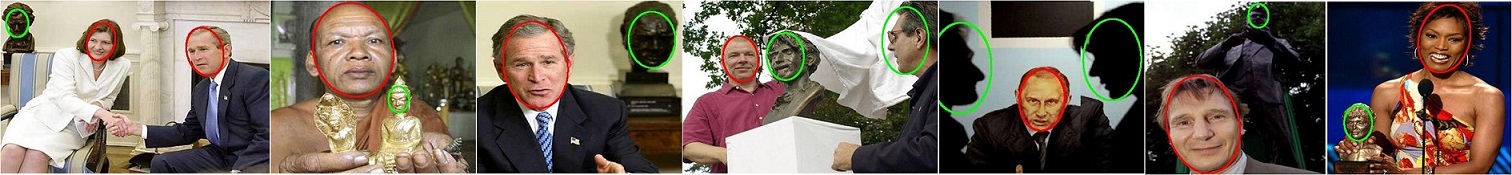}}
\subfigure[Miscellaneous faces]{
\includegraphics[width=1.0\textwidth]{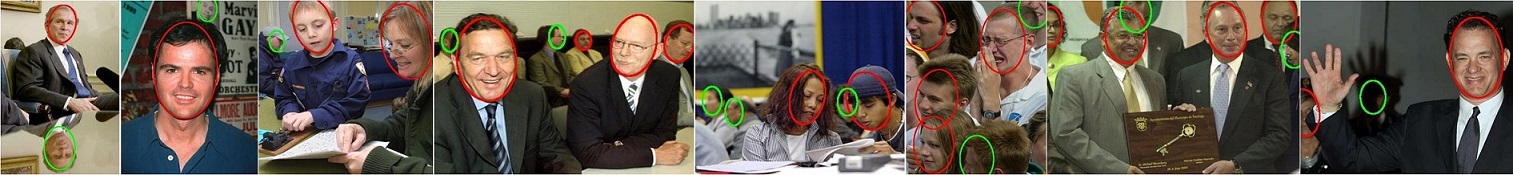}}
\caption{Examples of our manually labelled faces on the FDDB dataset. Red ellipses are the faces that FDDB has already labelled, green ellipses are the newly added faces.}\label{fig:sfaces}
\end{figure}

We add 238 unlabelled faces whose height and width are more than 20 pixels. Some examples are shown in Fig.~\ref{fig:sfaces}. 

\section{Ablative analysis of each detection layers}
To examine the contribution of each detection layers on the mAP performance, we progressively remove the detection layers to test their contribution on the WIDER FACE Val set. The detailed experiment results are listed in Tab.~\ref{tab:sablation}. After removing Conv3\_3 layer, the mAP changes are +$0.3\%$(Easy), +$0.5\%$(Medium) and -$24.7\%$(Hard), showing Conv3\_3 is crucial to detect small faces, but tiling plenty of smallest anchors also slightly hurts medium and large face detection performance. Besides, the most contribution of Easy and Medium subset are Conv5\_3 ($25.8\%$) and Conv4\_3 ($20.6\%$), respectively.

\vspace{-0.25cm}
\begin{table}[htbp]
\centering
\begin{tabular}{c|cccccc}
\toprule[2pt]
\multicolumn{1}{c|}{\bf Detection layers}&\multicolumn{6}{c}{\bf Ablative analysis}\\
\midrule[1pt]
Conv3\_3 & {\bf \texttimes} & & & & &\\
Conv4\_3 & & {\bf \texttimes} & & & &\\
Conv5\_3 & & & {\bf \texttimes} & & &\\
Conv\_fc7 & & & & {\bf \texttimes} & &\\
Conv6\_2 & & & & & {\bf \texttimes} &\\
Conv7\_2 & & & & & & {\bf \texttimes}\\
\midrule[1pt]
mAP changes on Easy subset (\%) & +0.3 & -0.6 & \bf{-25.8} & -10.2 & -3.2 & -1.4\\
mAP changes on Medium subset (\%) & +0.5 & \bf{-20.6} & -12.2 & -5.0 & -1.5 & -0.7\\
mAP changes on Hard subset (\%) & \bf{-24.7} & -8.7 & -4.1 & -1.8 & -0.6 & -0.2\\
\bottomrule[2pt]
\end{tabular}
\vspace{0.2cm}
\caption{The ablative results of each detection layers on the WIDER FACE Val set.}\label{tab:sablation}
\end{table}

\section{Latest results on the WIDER FACE dataset}
\begin{figure*}[htbp]
\centering
\subfigure[Val: Easy]{
\label{fig:sve} 
\includegraphics[width=0.325\textwidth]{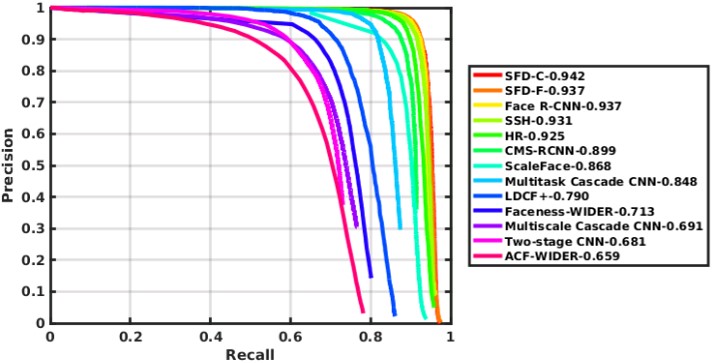}}
\subfigure[Val: Medium]{
\label{fig:svm} 
\includegraphics[width=0.325\textwidth]{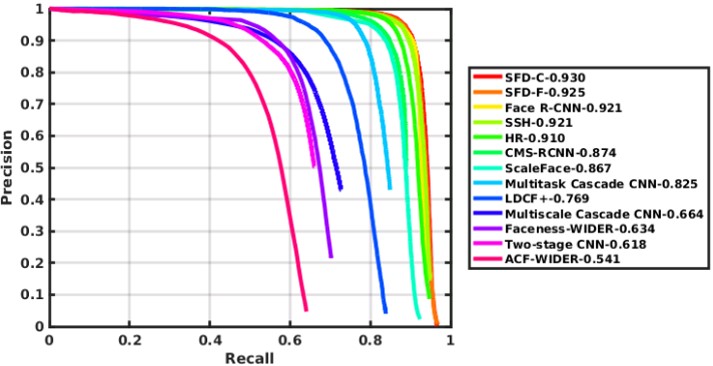}}
\subfigure[Val: Hard]{
\label{fig:svh} 
\includegraphics[width=0.325\textwidth]{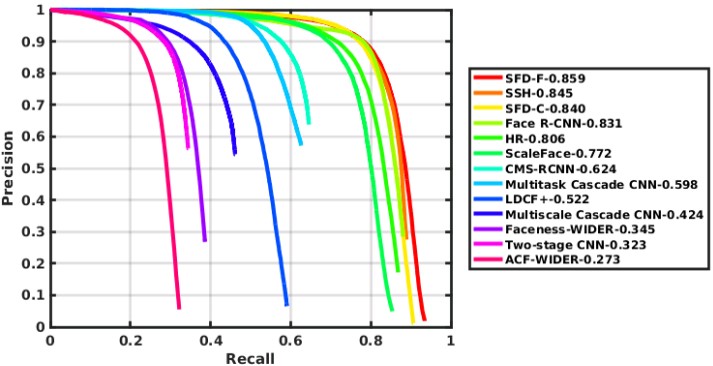}}
\subfigure[Test: Easy]{
\label{fig:ste} 
\includegraphics[width=0.325\textwidth]{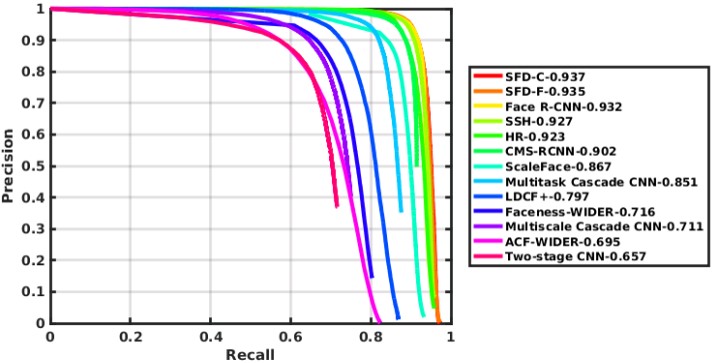}}
\subfigure[Test: Medium]{
\label{fig:stm} 
\includegraphics[width=0.325\textwidth]{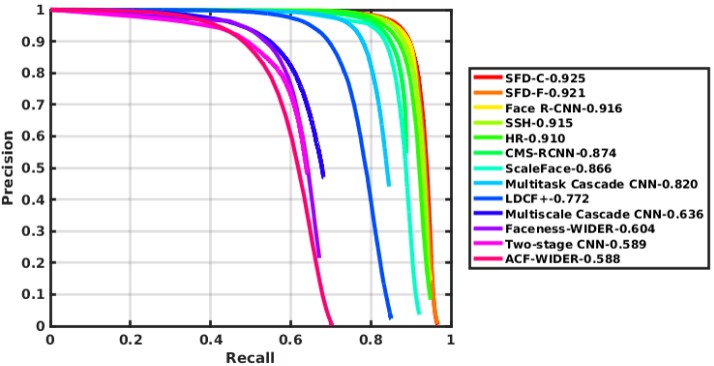}}
\subfigure[Test: Hard]{
\label{fig:sth} 
\includegraphics[width=0.325\textwidth]{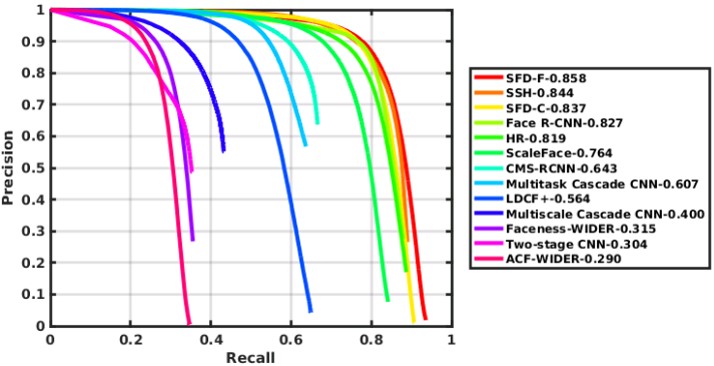}}
\caption{The latest precision-recall curves on WIDER FACE validation and test sets.\protect\footnotemark}
\label{fig:sWIDER_FACE} 
\end{figure*}

Fig.~\ref{fig:sWIDER_FACE} shows the latest precision-recall (PR) curves of our S$^3$FD (\textit{i.e.,} SFD-F and SFD-C) on WIDER FACE validation and test sets. SFD-F and SFD-C are our upgraded detectors. SFD-F further improves the detection ability of small faces and SFD-C focuses more on big and medium faces. The RP curves of SFD-F and SFD-C can be downloaded from the official website of WIDER FACE dataset\footnote{\url{http://mmlab.ie.cuhk.edu.hk/projects/WIDERFace/WiderFace_Results.html}}, which plots only the RP curves of SFD-F on its figure with the legend ``SFD". Our S$^3$FD achieves the best average precision on all subsets, i.e. $\textbf{0.942}$ (Easy), $\textbf{0.930}$ (Medium) and $\textbf{0.859}$ (Hard) for validation set, and $\textbf{0.937}$ (Easy), $\textbf{0.925}$ (Medium) and $\textbf{0.858}$ (Hard) for testing set.

\vspace{-0.2cm}
\footnotetext{Note-worthily, the latest evaluation code and annotation are used to generate these PR curves, while the results of WIDER FACE reported in our above paper are generated from the previous version of evaluation code or annotation.}

\end{document}